\documentclass[conference]{IEEEtran}
\IEEEoverridecommandlockouts
\usepackage{cite}
\usepackage{amsmath,amssymb,amsfonts}
\usepackage{algorithmic}
\usepackage{graphicx}
\usepackage{textcomp}
\usepackage{xcolor}
\usepackage{subfig} 
\usepackage{algorithm}
\def\BibTeX{{\rm B\kern-.05em{\sc i\kern-.025em b}\kern-.08em
    T\kern-.1667em\lower.7ex\hbox{E}\kern-.125emX}}
\renewcommand{\baselinestretch}{0.99}
\newtheorem{Remark}{Remark}

\begin{document}

\title{Multi-Modal Self-Supervised Semantic Communication\vspace{-2mm}\\
 }
 
\author{\IEEEauthorblockN{Hang Zhao, Hongru Li, Dongfang Xu, Shenghui Song, and Khaled B. Letaief}\vspace{-2mm}\\\IEEEauthorblockA{The Hong Kong University of Science and Technology, Hong Kong\\
Email: \{hzhaobi,\hspace*{1mm}hlidm\}@connect.ust.hk, \{eedxu,\hspace*{1mm}eeshsong,\hspace*{1mm}eekhaled\}@ust.hk.\vspace{-4mm}}
}

\maketitle

\begin{abstract}
Semantic communication is emerging as a promising paradigm that focuses on the extraction and transmission of semantic meanings using deep learning techniques. While current research primarily addresses the reduction of semantic communication overhead, it often overlooks the training phase, which can incur significant communication costs in dynamic wireless environments. To address this challenge, we propose a multi-modal semantic communication system that leverages multi-modal self-supervised learning to enhance task-agnostic feature extraction. The proposed approach employs self-supervised learning during the pre-training phase to extract task-agnostic semantic features, followed by supervised fine-tuning for downstream tasks. This dual-phase strategy effectively captures both modality-invariant and modality-specific features while minimizing training-related communication overhead. Experimental results on the NYU Depth V2 dataset demonstrate that the proposed method significantly reduces training-related communication overhead while maintaining or exceeding the performance of existing supervised learning approaches. The findings underscore the advantages of multi-modal self-supervised learning in semantic communication, paving the way for more efficient and scalable edge inference systems.
\end{abstract}


\section{Introduction}
Semantic communication is poised to play a pivotal role in sixth-generation (6G) wireless networks, particularly within the realms of the Internet of Things (IoT) and immersive extended reality (XR) systems. Unlike traditional data-oriented communication, which focuses on the transmission and reconstruction of data, semantic communication emphasizes the extraction and transmission of meaningful information, facilitated by deep learning techniques. In this paradigm, the transmitter extracts relevant semantic information from the input data tailored to the communication task, while the receiver interprets this information to accomplish the intended task. This approach can significantly reduce communication overhead and enhance system performance for specific tasks \cite{bourtsoulatze2019deep, xie2021deep, li2024tackling}.
\par
While previous studies have demonstrated the impressive performance of semantic communication, they have primarily focused on single-modal data tasks \cite{xie2024deep, han2022semantic}, which capture only a limited amount of information. In contrast, multi-modal data, due to its complementary nature, conveys richer semantic information \cite{luo2022multimodal}, making it essential to explore multi-modal semantic communication. Recent research has begun to investigate the integration of multiple data modalities. For example, the authors of \cite{xie2022task} proposed a transformer-based framework to unify the structure of transmitters for different tasks using multi-modal data. Similarly, the authors of \cite{wang2021deep} introduced a feature fusion method to address object detection and classification tasks using image and depth data.
\par
However, existing research has predominantly concentrated on minimizing communication overhead during inference, often neglecting the training phase. When tasks change, retraining the encoder and decoder necessitates significant gradient transmission, leading to high costs in terms of time and computational resources, particularly in multi-modal scenarios. To mitigate this, the authors of \cite{Li2025remote} suggested minimizing communication overhead during training by exchanging intermediate features and gradients over wireless channels. Nonetheless, research focused on multi-modal configurations remains limited, as the independent training of separate modalities often results in redundant transmissions across communication channels.
\par
To address this inefficiency, we propose a two-stage training approach for multi-modal semantic communication. The first stage employs a multi-modal self-supervised learning method, inspired by DeCUR \cite{wang2023decur}, to pre-train the transmitters. This approach leverages self-supervised learning to uncover inherent patterns in the data without requiring labels, facilitating task-agnostic and label-free pre-training. In the second stage, both the transmitter and receiver undergo task-specific fine-tuning through supervised learning. The pre-training stage enables the transmitter to learn the data distribution, thereby accelerating subsequent fine-tuning and reducing communication overhead. Simulation results demonstrate that the proposed method significantly reduces training-related communication overhead while achieving performance comparable to or even exceeding that of state-of-the-art supervised techniques.
\par
The paper is organized as follows. Section II presents the framework of semantic communication. Section III describes the proposed method, and Section IV presents the numerical results. Finally, Section VII concludes the paper.
\par
\textit{Notation:} Random variables and their corresponding realizations are denoted by boldface capital case (e.g., ${\boldsymbol{X}}$) and boldface lower letters (e.g., $\boldsymbol{x}$), respectively. $\mathbb{R}^{N\times M}$ and $\mathbb{C}^{N\times M}$ denote the space of $N\times M$ real-valued and complex-valued matrices, respectively. The mutual information between random variables $\boldsymbol{X}$ and $\boldsymbol{Y}$ is denoted by $I(\boldsymbol{X}; \boldsymbol{Y})$, while the conditional mutual information between $\boldsymbol{X}^1$ and $\boldsymbol{Y}$ given $\boldsymbol{X}^2$ is represented by $I(\boldsymbol{X}^1;\boldsymbol{Y}|\boldsymbol{X}^2)$. The tripartite mutual information is denoted by $I(\boldsymbol{X}^1;\boldsymbol{X}^2;\boldsymbol{Y})$. $||\cdot||^2$ denotes the $l_2$-norm of its argument, and the operator $\text{concat}(\cdot)$ represents the concatenation of the vectors. 
\section{Framework of multi-modal semantic communication system}
\begin{figure}[t]
    \centering
    \includegraphics[width=0.5\textwidth]{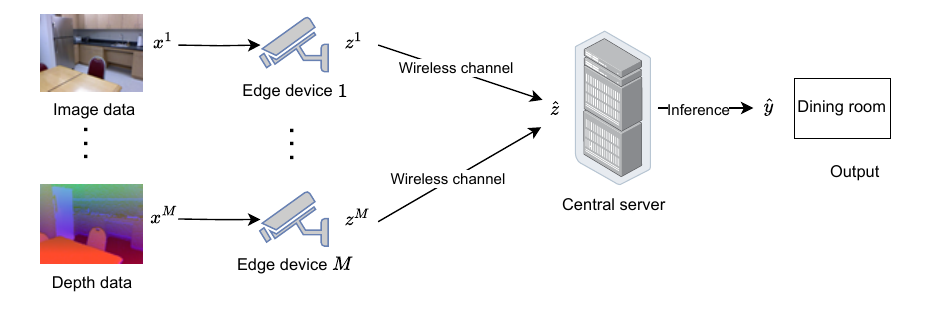} 
    \caption{Framework of the multi-modal communication system.}
    \label{framework}
\end{figure}
As illustrated in Fig. \ref{framework}, we consider a multi-modal semantic communication system for edge-device co-inference in classification tasks. In particular, we assume the presence of $M$ edge devices, each equipped with a sensor and an encoder to handle the data of one modality. At the $m$-th edge device, the sensor captures the raw data $\boldsymbol{x}^m \in\mathbb{R}^D$(with dimensionality $D$). The encoder then extracts features $\boldsymbol{z}^m \in\mathbb{R}^K$ (with dimensionality $K$) from the raw data $\boldsymbol{x}^m$ and transmits them to a central server via a wireless channel. Upon receiving the features $\hat{\boldsymbol{z}}^m \in\mathbb{C}^K $, the central server processes them and generates the inference results $\hat{\boldsymbol{y}}$. We collect the indices of the modalities in set $\mathcal{M} = \{1, \cdots, M\}$. 
\par
At the edge device, the raw data of each modality $\boldsymbol{x}^m $ 
 is input into the semantic encoder $f^m(\cdot; \boldsymbol{\alpha}^m)$ with parameters $\boldsymbol{\alpha}^m$ to extract semantic information, represented by:
\begin{equation}
\boldsymbol{z}^m = f^m(\boldsymbol{x}^m; \boldsymbol{\alpha}^m), \hspace*{2mm} m \in \mathcal{M}, 
\end{equation}
where $\boldsymbol{z}^m $ denotes the semantic feature for modality $m$.


The encoded vectors, $\boldsymbol{z}^m$, $\forall m \in \mathcal{M}$ , are then sent to the central server. The received vector for each modality is given by:
\begin{equation}
\hat{\boldsymbol{z}}^m = h^m  \boldsymbol{z}^m + \boldsymbol{n}^m, \hspace*{2mm} m \in \mathcal{M},
\end{equation}
where $h^m\in\mathbb{C}$ denotes the channel coefficient between the $m$-th edge device and the central sever, and $\boldsymbol{n}^m \sim \mathcal{CN}(\boldsymbol{0}, \sigma_{m}^2\boldsymbol{I})$ represents the additive white Gaussian noise (AWGN) with variance $\sigma_{m}^2$. 
The features $\hat{\boldsymbol{z}}^m $ for all modalities are concatenated at the central server to form the overall received feature:
\begin{equation}
\hat{\boldsymbol{z}} = \text{concat}(\hat{\boldsymbol{z}}^1, \hat{\boldsymbol{z}}^2, \cdots, \hat{\boldsymbol{z}}^M).
\end{equation}
Then, the server performs inference based on the concatenated received features $\hat{\boldsymbol{z}}$. In particular, the decoding process can be formulated as:

\begin{equation}
\hat{\boldsymbol{y}} = g(\hat{\boldsymbol{z}}; \boldsymbol{\phi}),
\end{equation}
where $g(\cdot;\boldsymbol{\phi})$ denotes the decoder at the receiver, and $\boldsymbol{\phi}$ represents the trainable parameters of the receiver.

While the above framework establishes the fundamental workflow of multi-modal transmission, practical deployment faces critical challenges in dynamic environments. Notably, in real-world scenarios where edge devices operate under volatile network conditions, two key constraints emerge:
\begin{enumerate}
\item \textbf{Delayed task awareness}: Label information and downstream tasks remain inaccessible prior to connection establishment.
\item \textbf{Training-phase overhead}: Frequent parameter updates during conventional end-to-end training incur prohibitive communication costs.
\end{enumerate}
These operational issues cause a fundamental tension between achieving high inference accuracy (requiring detailed feature transmission) and maintaining sustainable communication overhead during the training phase. The subsequent section addresses this challenge through a novel training methodology specifically designed for dynamic edge environments.
\section{Proposed self-supervised multi-modal semantic communication} 
\begin{figure*}[t]
    \centering
    \includegraphics[width=0.65\textwidth]{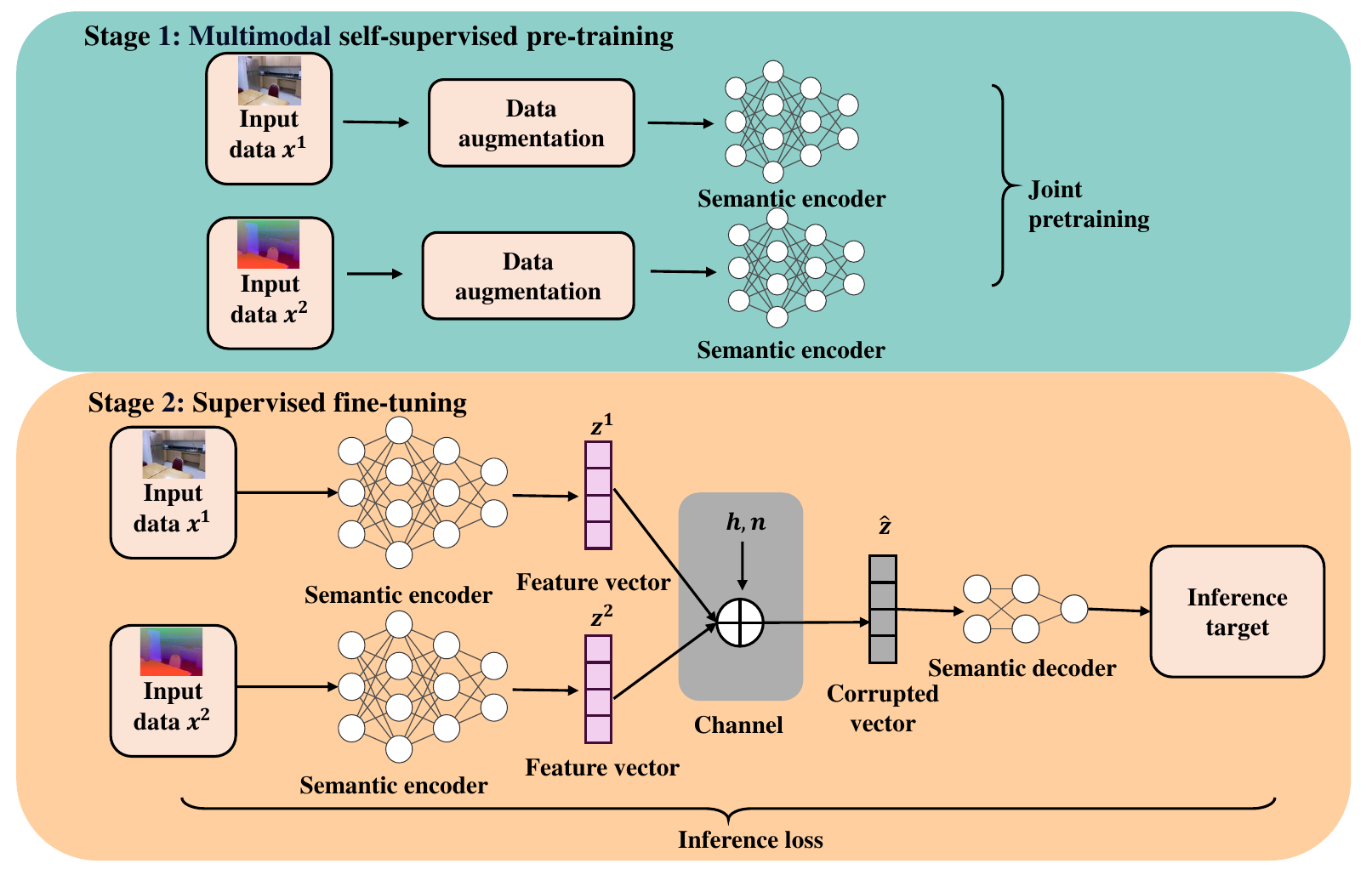} 
    \caption{Proposed multi-modal communication system that first conducts self-supervised encoder training, followed by end-to-end joint training. The system processes two modalities: RGB and depth data.}
    \label{system}
\end{figure*}
The objective of this work is to reduce the communication overhead in training stage for multi-modal semantic communication.  
As illustrated in Fig. \ref{system}, we propose an efficient self-supervised methodology comprising two steps: 1) multi-modal pre-training for task-agnostic feature learning at edge devices, and 2) joint task-specific fine-tuning for both edge devices and the central server. The first stage leverages the joint distribution of raw data from different edge devices to establish robust cross-modal representations without label dependency, effectively reducing communication costs through compressed semantic features. The second stage enables downstream task adaptation through supervised learning, where the edge devices and central server coordinate parameter updates using labeled data. For ease of illustration,  we consider the scenario with two modalities, e.g., RGB and depth data.
\subsection{Stage I : Multi-modal self-supervised pre-training}

In \cite{Li2025remote}, a single-modal self-supervised method was employed for pre-training. However, as will be demonstrated in subsequent analysis, this approach leads to performance degradation in multi-modal scenarios. To address this limitation, we propose a multi-modal self-supervised method that achieves superior performance. For that purpose, we begin with an information-theoretic analysis of key stages in the semantic communication framework during pre-training.

\begin{figure}[htbp]
    \centering
    \includegraphics[width=2.6in]{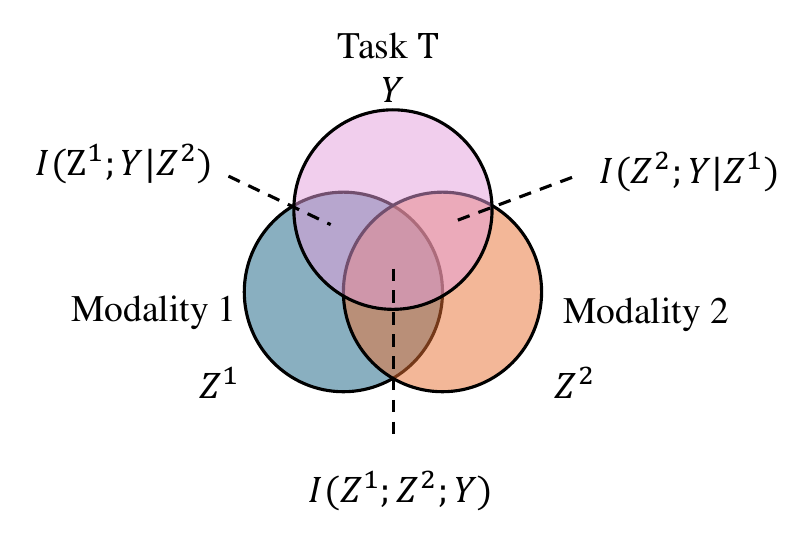} 
    \caption{Mutual information analysis for single-modal and multi-modal pre-training methods.}
    \label{venne}
\end{figure}
\subsubsection{Mutual Information in Pre-training}
Considering the case with two modalities, the Venn diagram in Fig. \ref{venne} quantifies the key advantages of multi-modal pre-training through the relationship between several information-theoretic measures, including: each modality's unique information $I(\boldsymbol{Z}^1;\boldsymbol{Y}|\boldsymbol{Z}^2)$ and $I(\boldsymbol{Z}^2;\boldsymbol{Y}|\boldsymbol{Z}^1)$, and the cross-modal information $I(\boldsymbol{Z}^1;\boldsymbol{Z}^2;\boldsymbol{Y})$. Here $\boldsymbol{Z}^1$, $\boldsymbol{Z}^2$ and $\boldsymbol{Y}$ denote the corresponding random variables of semantic features $\boldsymbol{z}^1$, $\boldsymbol{z}^2$ and label $\boldsymbol{y}$, respectively.
\par
Single-modal pre-training only captures the single-modal information $I(\boldsymbol{Z}^1;\boldsymbol{Y})$ and $I(\boldsymbol{Z}^2;\boldsymbol{Y})$, while multi-modal approaches jointly exploits both the unique information for each mode $I(\boldsymbol{Z}^2;\boldsymbol{Y}|\boldsymbol{Z}^1) $ and $I(\boldsymbol{Z}^1;\boldsymbol{Y}|\boldsymbol{Z}^2) $ and the shared cross-modal information $ I(\boldsymbol{Z}^1;\boldsymbol{Z}^2;\boldsymbol{Y}) $. This explains the performance degradation when using single-modal pre-training in multi-modal downstream tasks. 
\par
In single-modal scenarios, methods such as \cite{chen2020simple, zbontar2021barlow} prioritize maximizing task-relevant features within a single modality, expressed as \( I(\boldsymbol{Z}^1;\boldsymbol{Y}) \). In contrast, multi-modal scenarios require maximizing mutual information across all modalities:
\begin{equation}
\begin{aligned}
    I(\boldsymbol{Z}^1;\boldsymbol{Y}) + I(\boldsymbol{Z}^2;\boldsymbol{Y}) &= 2I(\boldsymbol{Z}^1;\boldsymbol{Z}^2;\boldsymbol{Y}) \\
    &\quad+ I(\boldsymbol{Z}^1;\boldsymbol{Y}|\boldsymbol{Z}^2) + I(\boldsymbol{Z}^2;\boldsymbol{Y}|\boldsymbol{Z}^1),
\end{aligned}
\end{equation}
\addtolength{\leftmargin}{0.02in}
which will lead to suppressing the unique characteristics related to the tasks of each modality (i.e., $I(\boldsymbol{Z}^1;\boldsymbol{Y}|\boldsymbol{Z}^2)$ and $I(\boldsymbol{Z}^2;\boldsymbol{Y}|\boldsymbol{Z}^1)$). This is because the weighting coefficients for the terms in the loss function prioritizes shared information $I(\boldsymbol{Z}^1;\boldsymbol{Z}^2;\boldsymbol{Y})$.
\par
To tackle the above mentioned issues, we propose a multi-modal framework that equally  maximizes both the shared and unique information:
\begin{equation}
\begin{aligned}
    I(\boldsymbol{Z}^1, \boldsymbol{Z}^2;\boldsymbol{Y}) &= I(\boldsymbol{Z}^1;\boldsymbol{Z}^2;\boldsymbol{Y}) \\
    &\hspace*{2mm} + I(\boldsymbol{Z}^1;\boldsymbol{Y}|\boldsymbol{Z}^2) + I(\boldsymbol{Z}^2;\boldsymbol{Y}|\boldsymbol{Z}^1).
\end{aligned}
\end{equation}
By comparing (6) with (5), we notice that these two objective functions differ in their main focuses. Specifically, single-modal methods overemphasize shared information in features while suppressing the unique feature relevant to each modality. On contrast, the proposed multi-modal pre-training method achieves comprehensive extraction of both the unique and shared information.
\subsubsection{Proposed Multi-modal Pre-training Approach}
It has been shown by \cite{wang2023decur} that decoupling representations integrate complementary information across different modalities,
which is critical for multi-modal pre-training. Based on the above analysis and inspired by \cite{wang2023decur}, we introduce a loss function that decouples cross-modal and intra-modal representations to learn both shared and unique information. Specifically, the overall training objective is given by:
\begin{equation}
    \mathcal{L}_{\text{pre-train}} = \mathcal{L}_\text{cross} + \mathcal{L}_\text{intra},
\end{equation}
where \( \mathcal{L}_{\text{cross}} \) and \( \mathcal{L}_{\text{intra}} \) denote the loss function for the cross-modal learning and intra-modal learning, respectively. To effectively learn these representations, we first apply data augmentation to the dataset \(\mathcal{D} = \{\boldsymbol{x}^m, \boldsymbol{y}\}_{m=1}^M\), resulting in augmented pairs \(\{\boldsymbol{x}^m, \tilde{\boldsymbol{x}}^m\}_{m=1}^M\). Then, we pass these augmented data through the encoder to obtain semantic features, yielding the feature sets \(\{\boldsymbol{z}^m, \tilde{\boldsymbol{z}}^m\}_{m=1}^M\).
\par
For each modality $m$ in  the intra-modal learning, we compute the intra-modal correlation matrix $\mathcal{C}^{m} \in \mathbb{R}^{K \times K}$ using augmented features:
\begin{equation}
    \mathcal{C}^{m}_{ij} = \frac{\sum_{b=1}^B z_{b,i}^m \tilde{z}_{b,j}^m}{\sqrt{\sum_{b=1}^B \left( z_{b,i}^m \right)^2} \sqrt{\sum_{b=1}^B \left( \tilde{z}_{b,j}^m \right)^2}},
\end{equation}
where $z_{b,j}^m $ denotes the $j$-th dimension of the original semantic feature in the $b$-th sample, and $\tilde{z}_{b,j}^m$ represents its \textit{augmented} counterpart via data transformation. $B$ is the batch size, $i,j \in [1,K]$ index feature dimensions, and $K$ represents the feature dimension size. Each entry $\mathcal{C}^{m}_{ij} \in [-1,1]$ quantifies the correlation between the $i$-th and $j$-th feature dimension of the $m$-th modality.
\par
Then, we define the intra-modal loss function $\mathcal{L}_{\text{intra}}$ as:
\begin{align}
    \mathcal{L}_{\text{intra}} &= \sum_{m=1}^M \mathcal{L}_{\text{intra}}^{m},
\end{align}
where $\mathcal{L}_{\text{intra}}^{m} $ represents the loss associated with the $m$-th modality, given by:
\begin{align}
    \mathcal{L}_{\text{intra}}^{m} &= \underbrace{\sum_{i=1}^K \left(1 - \mathcal{C}^{m}_{ii}\right)^2}_{\text{Diagonal alignment}} + \lambda_m \cdot \underbrace{\sum_{i=1}^K \sum_{\substack{j=1 \\ j \neq i}}^K \left( \mathcal{C}^{m}_{ij} \right)^2}_{\text{Off-diagonal decorrelation}},
\end{align}
with $\lambda_m > 0$ controlling the trade-off between feature stability (diagonal terms) and redundancy reduction (off-diagonal terms). The diagonal alignment terms ensure that each feature dimension remains consistent across data augmentations, while the off-diagonal decorrelation terms reduce redundancy by minimizing dependencies between different feature dimensions. 
\par
For cross-modal interaction between modality $m$ and $n$, we compute the \textbf{cross-modal correlation matrix} $\mathcal{C}^{m,n} \in \mathbb{R}^{K \times K}$ using features from different modalities as:
\begin{equation}
    \mathcal{C}^{m,n}_{ij} = \frac{\sum_{b=1}^B z_{b,i}^m z_{b,j}^n}{\sqrt{\sum_{b=1}^B \left( z_{b,i}^m \right)^2} \sqrt{\sum_{b=1}^B \left( z_{b,j}^n \right)^2}},
\end{equation}
which captures pairwise correlations between cross-modal feature dimensions.

To identify the shared and unique information between modalities, we decompose the correlation matrix into two distinct components:

\begin{itemize} \item \textbf{Shared Representation Block} ($\mathcal{C}_\text{sha}$): The shared block $\mathcal{C}^{m,n}[1:K_\text{sha}, 1:K_\text{sha}]$ captures the cross-modal features that are common across the modalities. \item \textbf{Unique Representation Block} ($\mathcal{C}_\text{uni}$): The unique block $\mathcal{C}^{m,n}[K_\text{sha}+1:K, K_\text{sha}+1:K]$ retains the modality-specific features that distinguish each modality. \end{itemize}
where $K_\text{sha} + K_\text{uni} = K$, ensuring the total size of the correlation matrix is preserved.

To effectively learn the shared and unique features across modalities, we introduce the cross-modal loss function: 
\begin{equation}
    \mathcal{L}_\text{cross} = \sum_{m=1}^M \sum_{n \neq m}^N \left( \mathcal{L}_\text{sha}^{m,n} + \mathcal{L}_\text{uni}^{m,n} \right),
\end{equation}
where the shared and unique loss components are given by:
\begin{alignat}{3}
    \hspace{0mm}\mathcal{L}_\text{sha}^{m,n} &= \underbrace{\sum_{i=1}^{K_\text{sha}} \left(1 - \mathcal{C}_{\text{sha},ii}^{m,n}\right)^2}_{\text{Shared feature alignment}} 
    &&\hspace*{-2mm}+ \lambda_\text{sha} 
    &&\cdot \hspace*{-4mm}\underbrace{\sum_{i \neq j} \left(\mathcal{C}_{\text{sha},ij}^{m,n}\right)^2}_{\text{Shared redundancy reduction}}, \\
    \hspace{0mm}\mathcal{L}_\text{uni}^{m,n} &= \underbrace{\sum_{i=1}^{K_\text{uni}} \left(\mathcal{C}_{\text{uni},ii}^{m,n}\right)^2}_{\text{Unique feature preservation}} 
    &&\hspace*{-2mm}+ \lambda_\text{uni} 
    &&\cdot \hspace*{-4mm}\underbrace{\sum_{i \neq j} \left(\mathcal{C}_{\text{uni},ij}^{m,n}\right)^2}_{\text{Unique feature decorrelation}},
\end{alignat}
with positive hypeparameters $\lambda_\text{sha}$ and $\lambda_\text{uni}$ controlling the trade-off between shared feature learning and unique feature disentanglement.
\par
with the given loss function, we can get the pre-trained encoders and decoder $\left\{\boldsymbol{\alpha}^{m\dagger}\right\}_{m=1}^M$: 
\begin{equation}
\left\{\boldsymbol{\alpha}^{m\dagger}\right\}_{m=1}^M = \mathop{\arg\min}\limits_{ \{\boldsymbol{\alpha}^{m}\}_{m=1}^M} \mathcal{L}_{\text{pre-train}}.
\end{equation}
\subsection{Stage II : Supervised fine-tuning}

In the second stage, we perform joint supervised training of the encoders and decoder. For that purpose, we define the cross entropy loss function:
\begin{equation}
\label{celossfun}
\mathcal{L}_{\text{fine}} = - \sum_i {y}_i \log(\hat{{y}}_i),
\end{equation}
where \( y_i \) represents the true label (or ground truth) for the \( i \)-th data point, and \( \hat{y}_i \) denotes the model-predicted probability to classify the same data point to the correct class. We apply the gradient descent method to obtain the optimal encoders and decoder 
$\left(\left\{\boldsymbol{\alpha}^{m*}\right\}_{m=1}^M, \boldsymbol{\phi}^*\right)$: 

\begin{equation}
\left(\left\{\boldsymbol{\alpha}^{m*}\right\}_{m=1}^M, \boldsymbol{\phi}^*\right) = \mathop{\arg\min}\limits_{ \{\boldsymbol{\alpha}^{m}\}_{m=1}^M, \boldsymbol{\phi}} \mathcal{L}_{\text{fine}}.
\end{equation}


\subsection{The Proposed Self-supervised Algorithm}
In earlier sections, we analyzed the feasibility of extracting shared and unique information from multi-modal data for task-agnostic feature learning. We also introduced a self-supervised approach based on mutual information maximization. In particular, during the pre-training stage, we use multi-modal contrastive learning method to train the encoders $\{f^m(\cdot; \boldsymbol{\alpha}^m)\}_{m=1}^M$, creating a universal model that captures the intrinsic structure of input data without relying on downstream task information. In the fine-tuning stage, we jointly update the pre-trained encoders $\{f^m(\cdot; \boldsymbol{\alpha}^m)\}_{m=1}^M$ on the device and the random inference model $g(\cdot;\boldsymbol{\phi})$ on the server using task-specific information. The proposed algorithm is outlined in Algorithm 1.
\begin{Remark}
In this paper, we focus on designing a deep learning-enabled pre-training and fine-tuning framework for semantic communication systems. The proposed framework can reliably support existing advanced resource allocation frameworks \cite{9832831}, \cite{zhou2025}, which facilitates efficient utilization of spectrum and power in semantic communication systems. For instance,  the authors of \cite{9832831} developed a reinforcement learning-based optimization framework to efficiently allocate wireless resource blocks for multiuser semantic communication systems. The authors of \cite{zhou2025} proposed a novel PODMAl framework enhanced by user demand prediction for spectrum resource allocation. The proposed method has achieved excellent performance and made a breakthrough for the spectrum decision-making optimization. 
\end{Remark}
\begin{algorithm}[t]
\caption{Multi-modal Self-Supervised Semantic Communication}
\label{alg:self_supervised}
\begin{algorithmic}
\STATE \textbf{Input:} Training data $\mathcal{D} = \{\boldsymbol{x}^m, \boldsymbol{y}\}_{m=1}^M$, batch size $B$, hyperparameters $\boldsymbol{\lambda}$, number of epochs $E_1$, $E_2$.
\STATE \textbf{Output:} Optimized model parameters $\{f^m(\cdot; \boldsymbol{\alpha}^m)\}_{m=1}^M$, $g(\cdot;\boldsymbol{\phi})$ for semantic communication.

\STATE \textbf{Joint Pre-training:}
\STATE \# No communication during pre-training.
\STATE Initialize $\{f^m(\cdot; \boldsymbol{\alpha}^m)\}_{m=1}^M$ with random weights.
\FOR{$e = 1$ to $E_1$}
    \STATE Sample a batch of data from $\mathcal{D}$.
    \STATE Apply data augmentation to obtain $\{\boldsymbol{x}^m, \tilde{\boldsymbol{x}}^m\}_{m=1}^M$.
    \STATE Compute the pre-training loss (7).
    \STATE Update the model parameters $\{f^m(\cdot; \boldsymbol{\alpha}^m)\}_{m=1}^M$ using gradient descent method.
\ENDFOR

\STATE \textbf{Fine-Tuning:}
\STATE Initialize fine-tuning model parameters $\{f^m(\cdot; \boldsymbol{\alpha}^{m\dagger})\}_{m=1}^M$ from pre-trained weights.
\FOR{$e = 1$ to $E_2$}
    \STATE Sample a batch of data from $\mathcal{D}$.
    \STATE Compute the fine-tuning loss (16).
    \STATE Update the model parameters $\{f^m(\cdot; \boldsymbol{\alpha}^m)\}_{m=1}^M$ and $g(\cdot;\boldsymbol{\phi})$ using gradient descent method.

\ENDFOR 
\end{algorithmic}
\end{algorithm}
\section{Numerical results}

\subsection{Experimental Setup}
To validate the effectiveness of the proposed algorithm, we test it over the  NYU Depth V2 (NYUD2) dataset, which contains 1,449 RGB-Depth image pairs spanning 27 indoor scene categories. Following the standard protocol \cite{han2022trusted}, we consolidate these categories into 10 distinct classes (9 primary scenes + 1 ``others'' category). The dataset is divided into 795 training samples and 654 test samples for experimental validation.

The proposed multi-modal encoders employ a ResNet-18 architecture \cite{he2016deep} for depth and image data, trained on a NVIDIA A40 GPU. Input images undergo resizing to $256 \times 256$ pixels followed by random cropping to $224 \times 224$ pixels for data augmentation. The batch size is set to 64, and the total training epochs
are 500. All training methods are performed using SGD  optimizer with weight decay and learning rate scheduling.

We track the progression of test accuracy for four methods over the training epochs. In the first three methods, the transmitter is pre-trained in a task-agnostic and label-free manner, followed by fine-tuning with downstream task and label information. In the fourth method, no pre-training is performed. A detailed description of each method is provided below:

\begin{itemize}
\item \textbf{SimCLR}: SimCLR\cite{chen2020simple} is a self-supervised learning framework that learns representations by maximizing the similarity between augmented views of the same image while minimizing the similarity between views of different images. The loss function is given by:
\begin{equation}
    \hspace*{-4mm}\mathcal{L} = - \log \left( \frac{\exp(\text{sim}(z_{i}, z_{j})/\tau)}{\sum_{k=1,k \neq i}^{2N} \exp(\text{sim}(z_{i}, z_{k})/\tau)} \right),
\end{equation}
where \( z_i \) and \( z_j \) represent embeddings of a positive pair (e.g., two augmented views of the same input sample), \( i \) and \( k \) index samples within a batch of size \( 2N \) (with each of the $N$ original samples), \( \text{sim}(\cdot, \cdot) \) denotes cosine similarity and \( \tau \) is temperature hyperparameter.

\item \textbf{Barlow Twins}: Barlow Twins\cite{zbontar2021barlow} is a self-supervised learning method that maximizes the cross-correlation matrix between representations of augmented views of the same image. The loss function is defined as:
\begin{equation}
    \mathcal{L}_{\text{Barlow Twins}} = \sum_i (1 - \mathcal{C}_{ii})^2 + \lambda \sum_i \sum_{j \neq i} \mathcal{C}_{ij}^2,
\end{equation}
where $\mathcal{C}$ is the cross-correlation matrix computed between the embeddings of two augmented views, $\mathcal{C}_{ij}$ denotes the $(i,j)$-th element of $\mathcal{C}$, and $\lambda$ is a weighting hyperparameter.
\item \textbf{Proposed}: The proposed effective multi-modal self-supervised learning method decouples common and unique representations across different modalities and enhances both intra-modal and cross-modal learning.
\item \textbf{Supervised}: The transmitters and receiver are trained from the start by supervised learning with the downstream task.
\end{itemize}



\begin{figure*}[t]
    \centering
    \captionsetup[subfigure]{labelformat=empty} 
    
    \subfloat[(a) Full labels, $\text{SNR}= 10$ dB]{
        \includegraphics[width=0.4\linewidth]{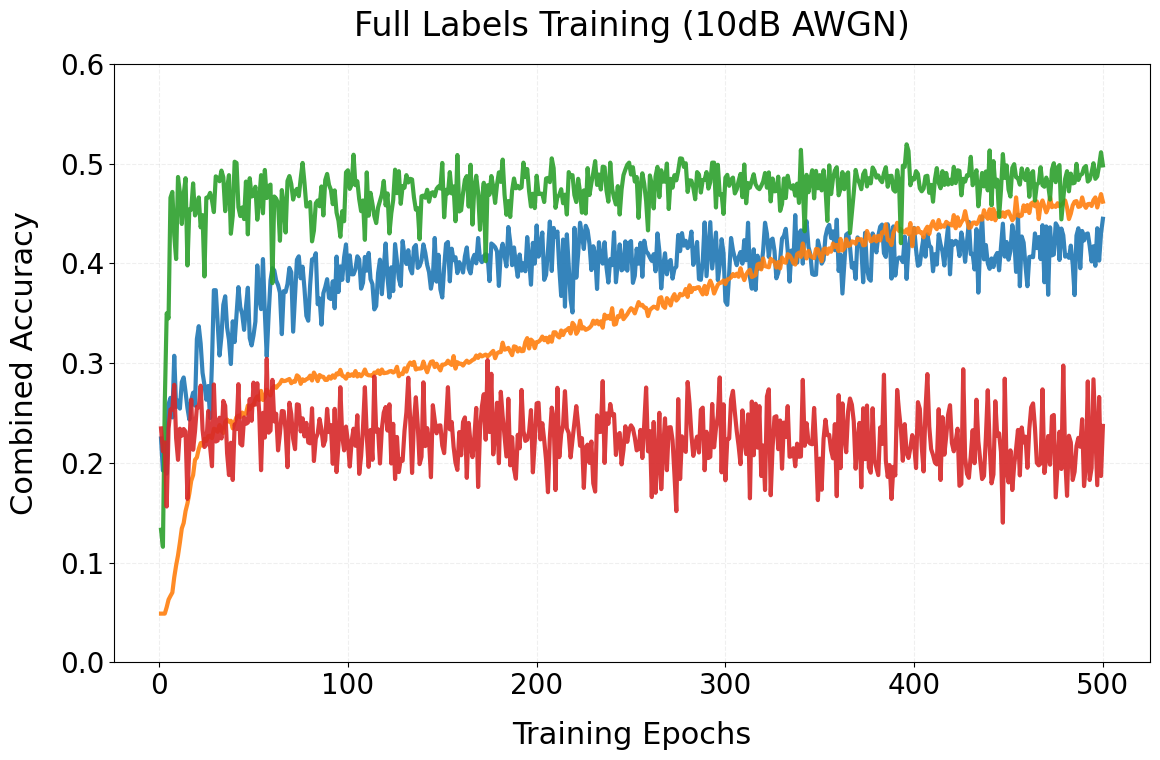}
    }\hspace{0.2cm} 
    \subfloat[(b) Full labels, $\text{SNR}= 20$ dB]{
        \includegraphics[width=0.4\linewidth]{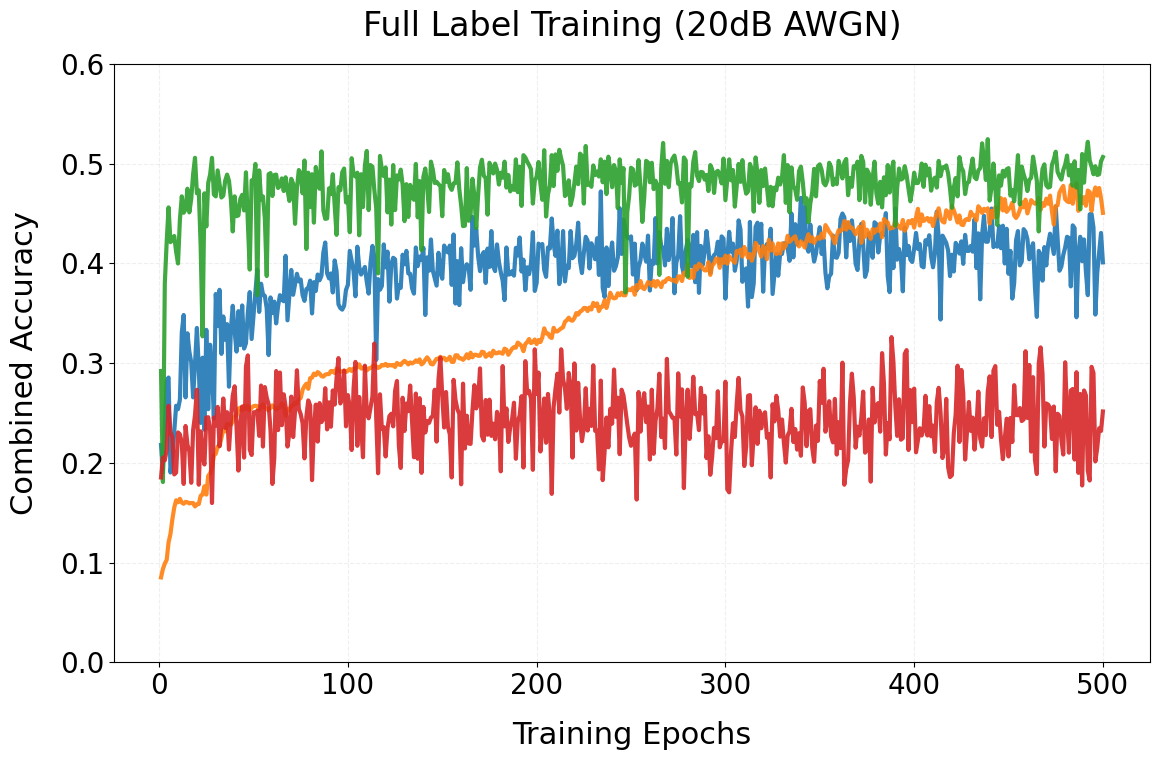}
    }
    \hspace{0.2cm}
    \subfloat[]{
        \centering
        \includegraphics[width=0.8\linewidth]{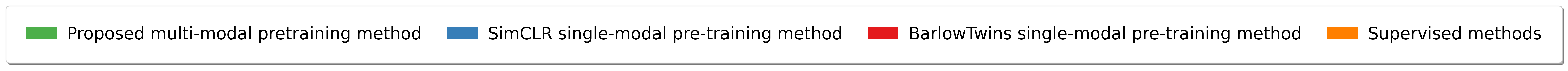}
    }
    \vspace{-2mm}
    \caption{The communication round versus test accuracy for the four methods with full labels.}
    \label{fig:results1}
\end{figure*}

In Fig. \ref{fig:results1}, we illustrate the test accuracy versus communication rounds for the four methods under different signal-to-noise (SNR) in AWGN channels. The proposed method exhibits significantly faster test accuracy improvement compared to fully supervised methods and consistently achieves better accuracy across all training stages and channel conditions. This improvement stems from the proposed methods' ability to capture task-relevant information during pre-training, which enables faster convergence during fine-tuning.

\begin{figure*}[t]
    \centering
    \captionsetup[subfigure]{labelformat=empty} 
    \subfloat[(a) Few labels, $\text{SNR}= 10$ dB]{
        \includegraphics[width=0.4\linewidth]{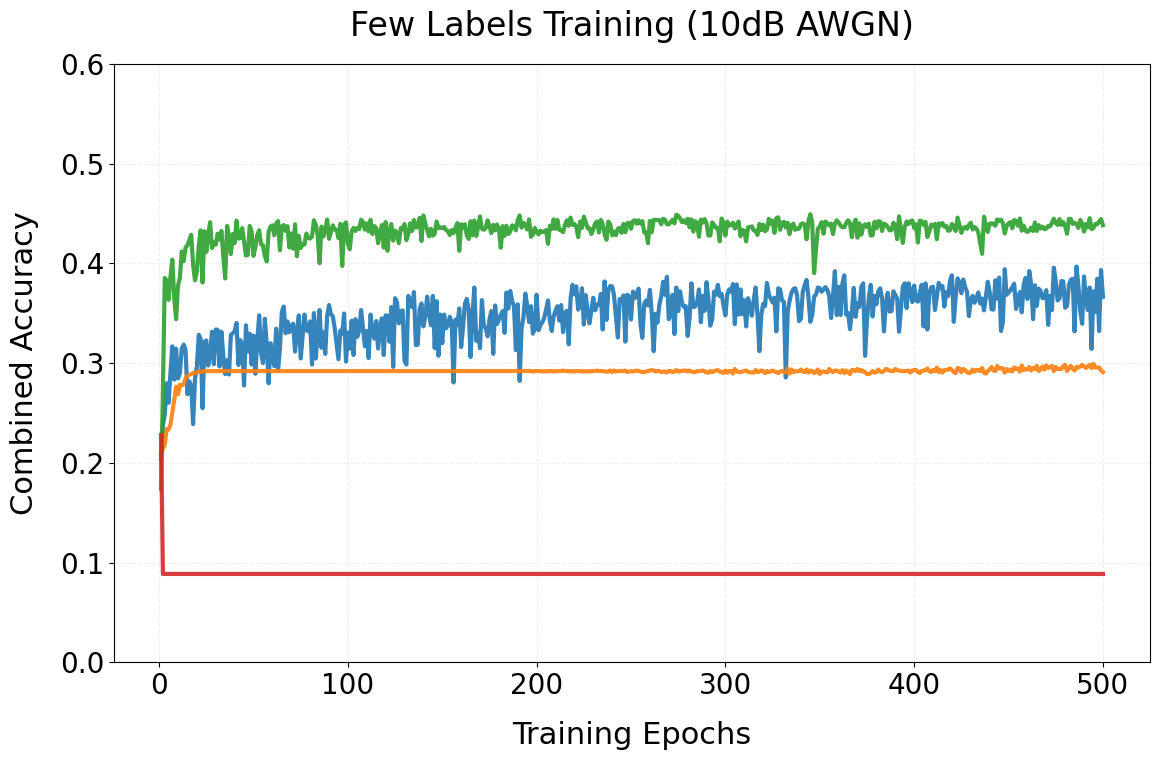}
    }\hspace{0.2cm}
    \subfloat[(b) Few labels, $\text{SNR}= 20$ dB]{
        \includegraphics[width=0.4\linewidth]{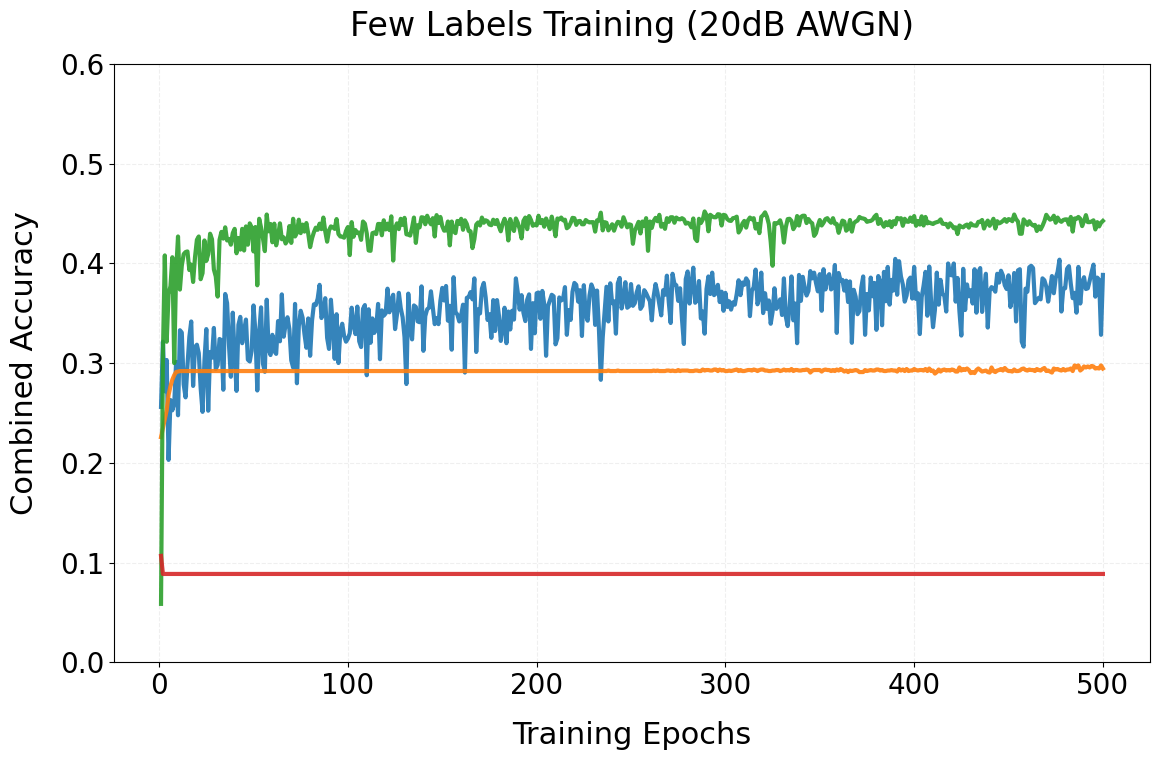}
    }
    \hspace{0.2cm}
    \subfloat[]{
        \centering
        \includegraphics[width=0.8\linewidth]{tag.png}
    }
    \vspace{-2mm}
    \caption{The communication round versus test accuracy for the four methods with 50\% labels.\vspace{-2mm}}
    \label{fig:results2}
\end{figure*}

As shown in Fig. \ref{fig:results2}, with only 50\% of the label information, the proposed method trained substantially outperforms the supervised baseline. This demonstrates the effectiveness of the proposed method in learning with limited labeled data. Notably, while the BarlowTwins pre-trained method suffers from exploding gradient issues under the same conditions, the proposed method maintains stable training dynamics, further highlighting its robustness in challenging scenarios.
    
    \vspace{0.2cm} 
\section{Conclusion}
\addtolength{\topmargin}{0.05in}
\addtolength{\rightmargin}{0.2cm}

This study presented a two-stage multi-modal semantic communication framework for edge inference, where edge devices transmit data from different modalities and share semantic features with an edge server for inference purposes. We proposed an efficient method that uses self-supervised learning to capture both shared and unique representations on the transmitter side, and then fine-tunes these representations for specific downstream tasks. Numerical results showed that the proposed method achieves excellent classification performance with fewer training rounds. Future research includes designing efficient coding schemes for dynamic network conditions and improving the system's scalability.

\bibliographystyle{IEEEtran}
\bibliography{reference}

\end{document}